\pdfoutput=1

\documentclass[11pt]{article}

\usepackage{acl}

\usepackage{times}
\usepackage{latexsym}

\usepackage[T1]{fontenc}

\usepackage[utf8]{inputenc}

\usepackage{microtype}

\usepackage{inconsolata}

\usepackage{graphicx}
\usepackage{placeins}

\usepackage{lipsum}
\usepackage{float}

\title{It Couldn't Help But Overhear: On the Limits of Modelling Meta-Communicative Grounding Acts with Supervised Learning}

\author{Brielen Madureira$^{\mathbf{1}}$ \hspace{10mm}  David Schlangen$^{\mathbf{1, 2}}$ \\
$^{\mathbf{1}}$Computational Linguistics, Department of Linguistics \\ University of Potsdam, Germany \\
$^{\mathbf{2}}$German Research Center for Artificial Intelligence (DFKI), Berlin, Germany \\
\texttt{\{madureiralasota,david.schlangen\}@uni-potsdam.de}}

\begin{document}
\maketitle

\begin{abstract}
    Active participation in a conversation is key to building common ground, since understanding is jointly tailored by producers and recipients. Overhearers are deprived of the privilege of performing grounding acts and can only conjecture about intended meanings. Still, data generation and annotation, modelling, training and evaluation of NLP dialogue models place reliance on the \textit{overhearing paradigm}. How much of the underlying grounding processes are thereby forfeited? As we show, there is evidence pointing to the impossibility of properly modelling human meta-communicative acts with data-driven  learning models. In this paper, we discuss this issue and provide a preliminary analysis on the variability of human decisions for requesting clarification. Most importantly, we wish to bring this topic back to the community's table, encouraging discussion on the consequences of having models designed to only ``listen in''.
\end{abstract}
\section{Is Grounding ``Supervisable''?}

``What are you looking at?'' asked Bob. ``Magpies are building a nest outside!'' Alice replied. If you were Bob, how would you continue that conversation? He could for instance say ``Awesome!'' or ``I saw that''. Whatever you say, it will probably differ from how he  continued: ``Building what?''. The decision to request clarification depends on mutual understanding, which is contingent on \textit{e.g.}~the current situation, the familiarity between interlocutors and the previous utterances. 
Or, more formally, it depends on the clarification potential of these utterances \citep{ginzburg2012interactive} and how they are assimilated into their \emph{common ground} \citep{clark1996cg}.

\begin{figure}[ht]
    \centering
    \fbox{\includegraphics[trim={0cm 18.6cm 16.3cm 0cm},clip,width=\columnwidth]{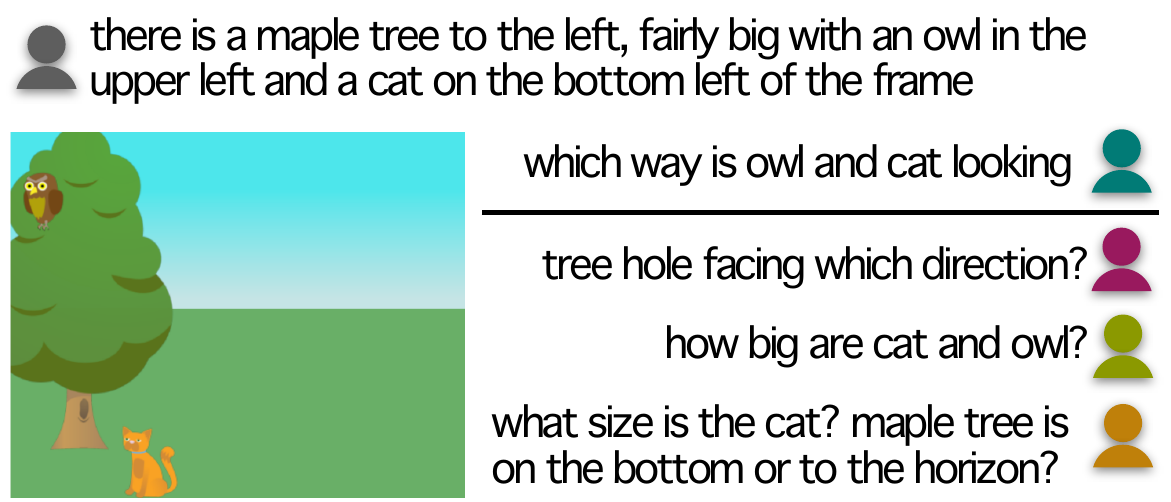}}
    \caption{Variability of clarification requests produced by three overhearers in comparison to the original one, in an instance of the instruction-following CoDraw dialogue game (\href{https://creativecommons.org/licenses/by-nc/4.0/}{CC BY-NC 4.0}), with cliparts from \citet{zitnick2013bringing}.}
    \label{fig:example}
\end{figure}

The one-to-many property of dialogue continuations is well-known in NLP \citep{zhao-etal-2017-learning,yeh-etal-2021-comprehensive,towle-zhou-2022-learn,liu-etal-2023-pvgru}. There is a combinatorial explosion of possibilities for any interaction \citep{bates-ayuso-1991-proposal,dingemanse2023interactive}, and individual human behaviour may vary at each point. This variability is hard to measure, since arguably no two people will ever be in the exact same situation with the same conversation history to react to \citep{yeomans2023practical}.

Still, the prevailing end-to-end deep learning methods commonly rely on supervised learning (SL) from a sample of human behaviour instantiating the reaction of \emph{a single human} at each observed context. Besides the issue of multiplicity of valid continuations, this paradigm faces another conceptual contention: dialogue models are trained to react upon a conversational history produced by someone else. In other words, they act as \textit{overhearers}\footnote{We will use this term to also mean reading or seeing signs. Also called \textit{observers} by \citet{georgila-etal-2020-predicting}.} of a dialogue in which they did not participate.

The suitability of data-driven methods and fixed corpora for modelling strategies and \emph{conversational} grounding phenomena like Clarification Requests (CR) has been questioned \citep{schatzmann-etal-2005-quantitative,benotti-blackburn-2021-recipe}. Static datasets of human observations have empirically failed to provide enough information to define a human-like CR policy \citep{testoni-fernandez-2024-asking,madureira-schlangen-2024-taking}. Moreover, chat-optimised LLMs mostly do not engage in grounding acts and, when they do, it does not fully align with human behaviour \citep{kuhn2022clam,deng-etal-2023-prompting,shaikh2023grounding}. The latter is not necessarily a problem: one can use other effective methods when it comes to building applications. But the first is: grounding is essential for human communication, and lack of it can lead to undesired breakdowns \citep{benotti-blackburn-2021-grounding}.

Since modelling human dialogue strategies and the use of meta-communivative acts remains an unsolved problem, we hereby wish to re-open the discussion on the consequences of overhearing, focusing on two grounding devices: backchannels and interactive repair \citep{fusaroli2017measures}.

\section{Overhearers in a Conversation}
\label{sec:lit}

As \citet{clark1996cg} defined it, in addition to speakers and addressees,\footnote{Or producers and recipients.} a conversation can have \emph{side-participants}, who are part of it but at a given moment are neither of the those two, and \emph{overhearers}, who are spectators without any rights or responsibilities, 
e.g.~a silent audience or a minute-taker who lacks the opportunity to interfere \citep{peters-2010-listening}. They are further divided into \emph{bystanders}, if one is aware of their presence, or \emph{eavesdroppers}, who listen secretly (or at a later time). There is evidence that the very process of understanding differs between addressees and overhearers: while interlocutors actively construct mutual understanding with each other, overhearers only passively consume the product of that process \citep{schober1989understanding}. 

Speakers can design their utterances while taking different attitudes towards overhearers when they are aware of their presence \citep{clark1992arenas,liu-etal-2016-coordinating}, but covert overhearers are not acknowledged at all in the conversation, and can only conjecture about the intended meanings \citep{clark1992arenas}. Although the grounding acts they witness, like backchannels, and the availability of multiple perspectives may indeed aid their comprehension \citep{tolins2016overhearers,tree2008overhearing}, the original interaction was opportunistically produced to be understood against the original participants' common ground \citep{schober1989understanding}.

In their corpus analysis of common ground in multi-party interactions, \citet{eshghi-healey-2007-collective} showed evidence that overhearers reach lower levels of understanding than ratified side participants, who in their turn are not very different from direct addressees, in what they call \emph{collective states of understanding}. Related to that, \citet{georgila-etal-2020-predicting} showed that observers and participants perceive interactions differently and the experiments by \citet{fox1999listening} provided evidence that overhearers can more easily comprehend instructions while listening to dialogues than to monologues. \citet{clark1992arenas} even argued that most psycholinguistic subjects are actually overhearers, so theories of language processing may actually be theories of overhearing, due to their lack of interactivity. 

Separating addressees from side participants and accommodating overhearers are salient problems in research on multi-party dialogue \citep{jovanovic-op-den-akker-2004-towards,ginzburg-fernandez-2005-scaling,traum-etal-2018-dialogue,parisse-etal-2022-multidimensional,ganesh-etal-2023-survey}.

\section{Are NLP Models Only Listening In?}
\label{sec:nlp}

More than a decade ago, \citet{rieser2011reinforcement} already discussed the limitations of using supervised approaches for learning dialogue strategies. They flagged up three concerns: textual data does not contain the underlying uncertainty measures, instances are treated as local point-wise estimates (instead of the sequences they really are) and exploration of novel strategies is not possible, since the model has access only to the outcomes of the chosen dialogue trajectory originally perpetrated by the humans. This reflects the (offline) \emph{overhearering paradigm}: a person or agent interpreting a pre-existing conversation and deciding what to do if they were in the original participants' shoes. 

In NLP, this paradigm is widely used in various modelling steps. Let us look closer at four main practices, which may have cascaded effects. 

\paragraph{Data Collection} Given the extra cost of coordinating the presence of more than one subject for generating dialogical data, especially in crowdsourcing campaigns, many strategies have been proposed to bypass that with overhearing. For instance, this happens when the data collection procedure is framed as a dialogue continuation task \citep{frommherz2021crowdsourcing}. To name a few related to grounding, we have \citet{zhou-etal-2022-reflect} who extracted dialogue contexts from existing datasets and presented them in a two-stage approach for some workers to generate common ground inferences and, separately, others generated a continuation as a response. Variations of overhearing manifest in techniques to generate CRs or their responses \citep{aliannejadi-etal-2021-building,gao2022dialfred,addlesee2024you} and are even embedded in data collection tools that allow dialogues to be constructed without persistent workers \citep{cascante-bonilla-etal-2019-chat}.

\paragraph{Annotation and Analysis} Corpus studies of interactive linguistic use can only be performed from an overhearer perspective, without full evidence of what participants intended and understood or the reasons for their decisions \citep{brennan-2000-invited,brennan2005conversation}. This is particularly challenging for research on common ground. For instance, \citet{rodriguez2004form} and \citet{schloder-fernandez-2015-clarifying} were confronted with the limitations of overhearers having only indirect access to the intentions of interlocutors when annotating CRs, partly remediating that by making a long dialogue context available. \citet{niekrasz-moore-2010-annotating} annotated references to conversation participants, joint actions that also serve to build common ground, emphasising that annotators were overhearers instructed to judge the speaker's intended purpose. Other annotations of grounding acts and common ground states had to rely on overhearers \citep{markowska-etal-2023-finding,zhang-etal-2023-groundialog,mohapatra2024conversational}.

\paragraph{Modelling} Prototypical data-driven models trained with supervised learning to \emph{process} dialogue, and possibly continue it, can, by design, be regarded as overhearers. This fact was made clear, for instance, in the CR model by \citet{schlangen-2004-causes}. \citet{traum2017computational} differentiated between the perspective of an observer in \emph{dialogue modelling} and of a participant in \emph{dialogue management}, stating that the main difference lies in the decision-making process of the latter, although some specific applications also exist for the first.

\paragraph{Evaluation} In human evaluation, overhearer experiments \citep{whittaker2005evaluating} are very common, even though it limits the judgements and measurements to user's \emph{perceptions} of the dialogue (rather than the actual behaviour) \citep{whittaker2005evaluating,foster2005assessing,moore-2011-language} and restricts assessment of metrics like effectiveness and efficiency \citep{paksima-etal-2009-evaluating}. It has historically been a ubiquitous approach due to advantages like having control on one aspect of the evaluation while avoiding navigational and timing aspects of real interactions \citep{villalba-etal-2017-generating}, avoiding interference from ASR and other technical problems \citep{buss-etal-2010-collaborating}, allowing the collection of feedback about alternative system responses \citep{walker2004generation} and avoiding natural language interpretation problems \citep{warnestal-etal-2007-emergent}. \citet{demberg-etal-2011-strategy} contrasted text overhearers with speech overhearers, pointing out that reading dialogues is artificially simplified, since participants can go back to difficult portions and choose the pace, and the two setups may also impact how evaluators rate the system. The available context may also have to be adjusted \citep{spanger-etal-2010-towards}. \citet{cercas-curry-rieser-2019-crowd} explicitly addressed the limitations of evaluation by overhearing and advocated for interaction with users. For a recent overview of works that use similar forms of \emph{static} evaluation, see \citep{finch-choi-2020-towards}.

As we have seen, the overhearing paradigm (fairly silently) permeates fundamental phases of dialogue modelling. The choice of this paradigm used to be a salient concept, with authors showing awareness of its limitations when it was employed.  \citet{kousidis2015power} even modelled a ratified side participant and had evaluators ``overhear the overhearer''. In recent publications, however, it is often taken for granted, as if it was the only natural way to go. What can be the consequences when it comes to cognitive models of conversational grounding?

\begin{figure*}[ht]
    \includegraphics[trim={0.65cm 0cm 4cm 0cm},clip,width=\textwidth]{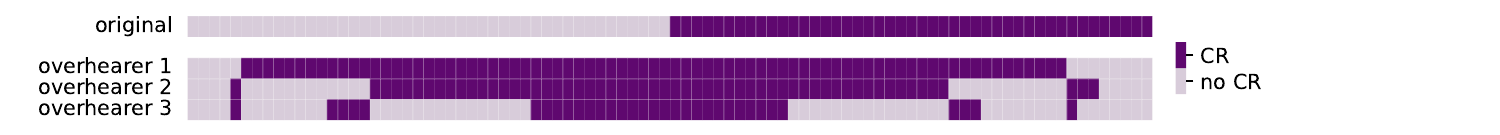}
    \caption{Variability in the decision of when to request clarification, comparing the decision of the original player with those of three overhearer annotators over 90 instances (horizontal axis) of the CoDraw game. Each cell is a data point and columns correspond to decisions on the same instance.}
    \label{fig:variation}
\end{figure*}
\section{Variability in Human Grounding Acts}

As humans speak, they can provide positive and negative evidence of mutual understanding \citep{clark1990grounding,roque-traum-2008-degrees}, but modelling their timing and decision-making is challenging. \citet{traum2017computational} claimed that ``it can be very difficult to efficiently capture regularities in behavioral patterns that lead to similar, but not identical structures''. In connection to that, people may take various paths in similar conversational situations \citep{bates-ayuso-1991-proposal}. It is thus an open question how far data-driven supervised learning can get given the inherent variability of explicit (not to mention the latent) collateral signs of grounding.

Backchannels, a positive evidence of grounding, 
were demonstrated to involve individual variability, and even idiosyncrasy, possibly due to personality
, gender or randomness \citep{huang2012crowdsourcing,backchannel-idiosyncratic}. Although those works showed some regularity in their timing, the SotA for the backchannel prediction task is not very high (.66 weighted F1) 
\citep{liermann-etal-2023-dialogue}.

Findings on the variability of human decisions to initiate a CR, a negative sign of grounding, are still sparse. \citet{stoyanchev-etal-2013-modelling} measured an absolute agreement of 39\% among three annotators for \emph{scripted} dialogues with missing ASR information. As another reference, \citet{shaikh2023grounding} reported a Cohen's $\kappa$ of 48.45 for clarification in emotional support conversations, which, they claimed, may even be inflated. The task of deciding when to request clarification in collaborative instruction following is under active investigation, but models' performance is still suboptimal \citep{shi-etal-2022-learning,li-etal-2023-python,madureira-schlangen-2023-instruction,mohanty2023transforming}. Recent works on the multimodal CoDraw dialogue game \citep{kim-etal-2019-codraw} argued that this may be due to the variability in human decisions and the limitations of using supervised learning \citep{testoni-fernandez-2024-asking,madureira-schlangen-2024-taking}.

\section{A Brief Analysis of Regularity in CRs}

In CoDraw, an instruction follower receives instructions to reconstruct a scene using cliparts (as in Figure \ref{fig:example}). Only the instruction giver sees the target scene. \citet{madureira-schlangen-2023-instruction} identified  all CRs (around 11\% of the instruction follower's utterances) and defined the task of deciding when to request clarification, where models reached only up to~.41 binary F1. What is missing as evidence for the claim that data-driven models cannot fully succeed in learning a ``when policy'' from human data is the actual human performance on this NLP task, i.e.~what \emph{overhearers} predict.

For an initial analysis, we collected a convenience sample with three annotators performing a similar task as the trained models: given a dialogue history and the current state of the reconstructed scene, decide which actions to take and, if needed, request clarification (details in Appendix). We randomly selected a sample with 90 instances; in half of them, the original player had produced a CR. 

The average binary F1 of overhearers with respect to the original decision was~.51, not much above what SotA models achieve. But the proportion of CRs widely ranged from 36 to 85\%. Among the three annotators, the Krippendorff's $\alpha$ was  $0.10$ and the mean pairwise Cohen's $\kappa$ was $0.18$. That is already low, but if we consider the original decision as a fourth annotator, measures are even lower: $\alpha$ was 0.02 and $\kappa$ was 0.06. This indicates that there was slightly more agreement among overhearers than among addresses and overhearers, but in general there was little agreement on deciding when a CR should be realised. Figure \ref{fig:variation} presents the main binary decision (whether to request clarification or not) for each of the 90 annotation instances, serving to provide a visual overview of such variability. 

In terms of surface forms, the average BLEU score was $0.11$ (std$=0.10$) using the original CR as a source and the produced utterances as a reference. The mean cosine similarity between the embedding of the produced and the original CRs was $0.38$, $0.29$ and $0.36$ for the three overhearers. Figure \ref{fig:example} shows an example of how diverse the produced clarifications can be, both in form and in content, even when all subjects made the same decision to clarify at a given point. 

These are preliminary insights from a pilot study. Further standardised experiments with a larger sample must be conducted. Still, the results are already useful to strengthen the argument that, like backchannelling, human CR decisions lack regularity and overhearers have a much harder task trying to interpret and act upon someone else's grounding acts. Decisions depend on how interlocutors distribute grounding costs, as per the principle of least collaborative effort \citep{clark1990grounding}. Besides, there might be adaptive behaviours that models are not capturing \citep{dideriksen2023quantifying}.

To continue this investigation, we propose distinguishing between the clarification potential \citep{ginzburg2012interactive,benotti-2009-clarification} and the clarification need. The first is a larger set of possibilities for clarification of a given utterance, while the latter refers to the decision of whether and what to clarify taken by a given individual operating with that utterance and identifying something worth clarifying. Or, in other words, the clarification need, which is \textit{in the agent}, refers to what was asked among all that could be asked. It is challenging to design experiments that can capture the clarification need among individuals, in particular due to the difficulty in replicating a given dialogue context for different subjects if they are not acting as overhearers. A possible next step is to turn the CR decision into an acceptability task, regarding it as a contrast. For each instance, the annotator would see a set of CRs. The actual CR observed in the data should ideally be accepted, but possibly others too. If the original CR falls into the empirical potential, there should be a plausible need for it at that point. Such experiment could also aim to measure uncertainty at each turn.

\section{Discussion}

Mutual understanding is crafted by ``interacting minds'' \citep{dingemanse2023beyond}. In dialogue, ``interlocutors share or synchronise aspects of their private mental states and act together in the world'' \citep{brennan2010two}. On the other hand, we have shown that the current NLP methodology mostly limits us to learning how \emph{overhearers} predict discourse representations without the actual joint decision making facet, due to the way that data is produced and annotated, the assumptions behind training mechanisms and the evaluation protocols, each adding a layer of overhearing.

What can be a better setup to learn human dialogue behaviour, realising it as a truly interactive process?  One needs to move on from one-off supervised learning to sequential models that not only \emph{understand} dialogues but also \emph{participate} in them.\footnote{See \citep{min-etal-2022-dont} for a related discussion on the limitations of imitation learning and behaviour cloning for embodied agents. See also \citep{ortega2021shaking} for a discussion on supervised learning and the sequential aspect of an interaction.} Reinforcement learning provides that framing with a fully accessible and explorable environment \citep{rieser2011reinforcement}, but somewhat circularly requires a good simulation of an user or interlocutor \citep{schatzmann-etal-2005-quantitative,georgila2006user,li-etal-2020-rethinking}. Although LLMs can serve as speaker simulators, so far they cannot fully model all dialogue phenomena. Another possibility are hybrid combinations of supervised and reinforcement learning \citep{henderson-etal-2008-hybrid}, as well as further improvements in techniques like RLHF, PPO and DPO. But independently of the learning regime, data-driven approaches, which rely on extracting latent patterns and regularities in a corpus, stumble upon the individual variability of some dialogue phenomena, so that tasks may be ill-defined in datasets. Besides, although transcribed dialogue contain clues about the decision making during a conversation, they provide only limited evidence of what participants understood or intended, or their internal states \citep{brennan2005conversation}, which are pertinent for modelling some dialogue decisions and meta-communicative acts.

Indeed, interfaces do not necessarily have to conform to human behaviour, as long as they can sustain \textit{graceful interaction} \citep{hayes-1980-expanding}. But from a cognitive perspective, the current NLP resources do not seem to satisfactorily meet our needs for modelling grounding mechanisms. To study the human mind, do we want cognitive models of how meaning and common ground are constructed or only of how they can be reverse engineered from someone else's interactions?

\paragraph{To conclude} With this argumentative paper, we wish to encourage more studies on the variability of human grounding acts and its impact in modelling human dialogue strategies. Besides, we advocate making the overhearing paradigm explicit whenever it is used in future publications and discussing how it can have influenced reported findings.

\section*{Acknowledgements}
We thank the anonymous reviewers for their valuable feedback. We are also thankful to the three annotators who took part on the pilot study.

\bibliography{anthology,custom}

\appendix
\section{Appendix}
\label{sec:appendix}


\paragraph{Annotation Task} The decisions from the overhearer perspective were performed by 3 annotators. Two of them are student assistants employed in our lab and one is a volunteer acquainted with the first author. A simple GUI interface showed the dialogue history (from 1 to 3 turns) up to the last instruction giver instruction, the current state of the reconstructed scene and the gallery of available cliparts. They could select up to 4 high level, discrete actions (add, move, resize, flip, delete) and the corresponding cliparts from dropdown lists. Besides, they could type a clarification request to continue the dialogue if they wished (otherwise, the next utterance field should be left blank). In future studies, a full interface similar to the original game should be used, i.e.~giving the opportunity for cliparts to be moved around and edited in the scene. Here, the selection of actions was just used to enforce that the overhearers reflected on the pertinent actions while deciding whether to request clarification. Note that the step of action taking makes annotators more privileged than plain overhearers that just process the dialogue, but it better approximates the decision of the iCR-Action-Taker models in \citet{madureira-schlangen-2024-taking}. In this case, they are overhearers of the dialogue context, but try to minimally act as a player doing the next step. The results work as an upper bound for plain overhearers.

\paragraph{Additional Details} The inter-annotator agreement metrics were computed with \texttt{nltk} using \texttt{chencherry.method3} for smoothing. The sentence embeddings for the CR utterances were computed with model \texttt{sentence-transformers/all-MiniLM-L6-v2} from SentenceTransformers \citep{reimers-gurevych-2019-sentence}.

\clearpage

\end{document}